%% file: main.tex
\documentclass[conference]{IEEEtran}
\usepackage{times}

\usepackage[numbers]{natbib}
\usepackage{multicol}
\usepackage[bookmarks=true]{hyperref}
\usepackage{graphicx}
\usepackage{subcaption}
\usepackage{booktabs}
\usepackage{amsmath,amsfonts}
\usepackage[ruled,vlined]{algorithm2e}
\usepackage{multirow}
\usepackage{booktabs}

\usepackage{caption}

\newcommand{\insertfig}{
  \begin{center}
    \includegraphics[width=\textwidth]{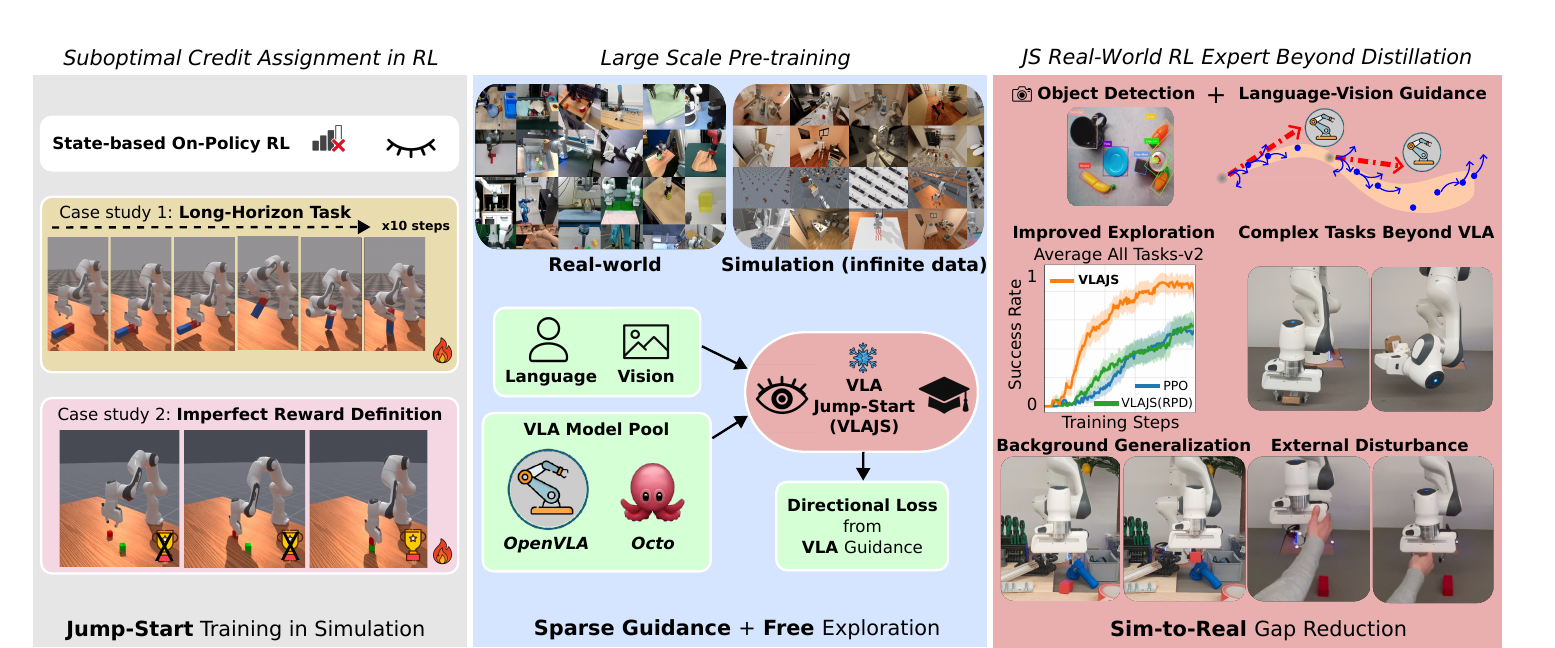}
    \captionof{figure}{\textbf{Overview of Vision-Language-Action Jump-Starting (VLAJS).} The figure illustrates the motivation, method, and outcomes of VLAJS. \textbf{Left:} We highlight suboptimal credit assignment in state-based, on-policy RL, focusing on: long-horizon tasks with extended action sequences and environments with imperfect reward design. \textbf{Center:} VLAJS leverages large-scale VLA pretraining from both real-world and simulation data. A pool of pretrained VLA models (\textit{e.g.}, OpenVLA) provides sparse, low-frequency action suggestions conditioned on language and vision. These suggestions are incorporated during RL training through a directional action-consistency loss, enabling jump-start learning while preserving exploration. \textbf{Right:} The VLAJS-based trained RL agent surpasses distillation-based approaches, achieving improved exploration, robustness to background changes and external disturbances, and reduced sim-to-real gaps.}
    \label{fig:simtoreal}
  \end{center}
}

\makeatletter
\apptocmd{\@maketitle}{\centering\insertfig\vspace{-1.5em}\setcounter{figure}{1}}{}{}

\makeatother

\begin{document}

%% paper title
\title{Vision-Language-Action Jump-Starting for
Reinforcement Learning Robotic Agents}

%% You will get a Paper-ID when submitting a pdf file to the conference system
%\author{Author Names Omitted for Anonymous Review.}

%% authors and affiliations
\renewcommand\IEEEauthorrefmark[1]{\textsuperscript{#1}}

\author{
\IEEEauthorblockN{Angelo Moroncelli\IEEEauthorrefmark{1}\IEEEauthorrefmark{2},
Roberto Zanetti\IEEEauthorrefmark{3},
Marco Maccarini\IEEEauthorrefmark{1}\IEEEauthorrefmark{2},
Loris Roveda\IEEEauthorrefmark{1}\IEEEauthorrefmark{3}}

\IEEEauthorblockA{\IEEEauthorrefmark{1}
University of Applied Science and Arts of Southern Switzerland, Department of Innovative Technologies,\\ IDSIA-SUPSI, Lugano, Switzerland \{angelo.moroncelli, marco.maccarini, loris.roveda\}@supsi.ch}

\IEEEauthorblockA{\IEEEauthorrefmark{2}
Università della Svizzera Italiana, Faculty of Informatics, Lugano, Switzerland}

\IEEEauthorblockA{\IEEEauthorrefmark{3}
Politecnico di Milano, Mechanical Department, Milano, Italy\\
\{roberto2.zanetti@mail., loris.roveda@\}polimi.it}
}

\maketitle

\begin{abstract}
Reinforcement learning (RL) enables high-frequency, closed-loop control for robotic manipulation, but scaling to long-horizon tasks with sparse or imperfect rewards remains difficult due to inefficient exploration and poor credit assignment. Vision-Language-Action (VLA) models leverage large-scale multimodal pretraining to provide generalist, task-level reasoning, but current limitations hinder their direct use in fast and precise manipulation. In this paper, we propose \emph{Vision-Language-Action Jump-Starting (VLAJS)}, a method that bridges sparse VLA guidance with on-policy RL to improve exploration and learning efficiency. VLAJS treats VLAs as transient sources of high-level action suggestions that bias early exploration and improve credit assignment, while preserving the high-frequency, state-based control of RL. Our approach augments Proximal Policy Optimization (PPO) with a directional action-consistency regularization that softly aligns the RL agent’s actions with VLA guidance during early training, without enforcing strict imitation, requiring demonstrations, or relying on continuous teacher queries. VLA guidance is applied sparsely and annealed over time, allowing the agent to adapt online and ultimately surpass the guiding policy. We evaluate VLAJS on six challenging manipulation tasks---lifting, pick-and-place, peg reorientation, peg insertion, poking, and pushing---in simulation, and validate a subset on a real Franka Panda robot. VLAJS consistently outperforms PPO and distillation-style baselines in sample efficiency, reducing required environment interactions by over 50\% in several tasks. Real-world experiments demonstrate zero-shot sim-to-real transfer and robust execution under clutter, object variation, and external perturbations.
\end{abstract}

%\IEEEpeerreviewmaketitle

\section{Introduction}
\label{sec:introduction}

Reinforcement Learning (RL) provides a powerful framework for learning closed-loop control policies directly from interaction~\citep{712192}. In robotics, RL enables high-frequency, state-based controllers that exploit rich proprioceptive and geometric feedback, enabling precise motor behaviors and online adaptation. These properties make RL particularly attractive for real-world manipulation, where robustness to disturbances, tight feedback loops, and reliability are critical~\citep{hil_serl_science}.

Despite these strengths, RL faces well-known challenges. Learning complex manipulation behaviors often requires long training times, careful reward engineering, and large amounts of interaction data~\citep{andrychowicz2021matters,liang2018gpu,taomaniskill3}. These issues are especially pronounced in long-horizon or suboptimally rewarded tasks, where delayed rewards lead to weak credit assignment and slow policy improvement~\citep{pignatellisurvey}.

Recent advances in Large Language Models (LLMs) and Vision-Language Models (VLMs) offer a complementary direction~\citep{hu2023generalpurposerobotsfoundationmodels,firoozi2023foundation,driess2023palme,ahn2022saycan}. Vision-Language-Action (VLA) models unify perception, language understanding, and control by mapping multimodal observations directly to robot actions~\citep{brohan2022rt1,brohan2023rt2}. Large-scale systems such as Octo~\citep{octo2024}, OpenVLA~\citep{kim2024openvla}, $\pi_0$~\citep{intelligence2025pi_}, and RDT-1B~\citep{liu2024rdt1b} demonstrate strong semantic understanding and broad generalization, often exhibiting zero-shot and few-shot capabilities.

However, current VLA models are not designed to replace RL-based controllers for precise robotic manipulation. Their reliance on large expert datasets limits scalability and adaptation~\citep{openx2023,zhao2023learning,khazatsky2024droid,nair2022learning}, while transformer-based architectures typically operate at low control frequencies, restricting their ability to handle precision requirements, disturbances, and long-horizon closed-loop behaviors~\citep{Moroncelli2024TheDO}. As a result, despite strong semantic priors, standalone VLA performance in real-world manipulation remains limited.

In this work, we view RL and VLA models as complementary components for training robotic manipulation policies. RL provides a deployable, high-frequency control backbone, while VLA models offer sparse, high-level action priors that encode semantic knowledge and task structure. Rather than contrasting these approaches, we ask:

\emph{Can the semantic knowledge captured by VLA models be used to accelerate RL, while preserving the precision, adaptability, and reliability of RL-based controllers?}

\begin{figure}[t]
    \centering
    \includegraphics[width=0.7\linewidth]{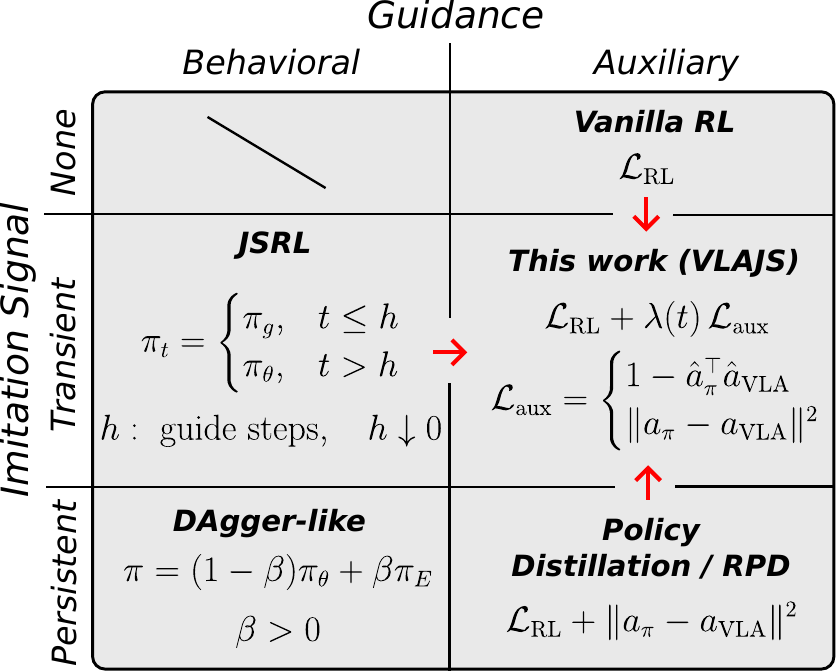}
    \caption{\textbf{Comparison of guidance strategies in RL.} Methods are categorized by guidance type (behavioral vs. auxiliary) and imitation persistence (none, transient, and persistent). Vanilla RL uses no guidance, DAgger-like methods apply persistent behavioral imitation, and policy distillation/RPD rely on persistent auxiliary losses. JSRL provides transient behavioral guidance, while \textbf{VLAJS} introduces transient auxiliary guidance via a directional action-consistency loss, accelerating early learning without limiting asymptotic performance.}
    \label{fig:sota}
\end{figure}

Fig.~\ref{fig:simtoreal} illustrates our perspective. We treat VLA models as sources of sparse, low-rate guidance that bias early exploration, improve credit assignment, and reduce the sim-to-real gap during training. The RL agent remains an on-policy, state-based learner operating at high control frequency, capable of exploiting precise feedback and ultimately surpassing the guiding policy.

To realize this synergy, we introduce \emph{VLAJS}, a method for jump-starting on-policy RL using sparse guidance from VLA models. A high-frequency RL controller receives occasional VLA action suggestions during early training, incorporated through a directional action-consistency regularization within a PPO framework. This guidance is transient: it is queried infrequently, temporally propagated across control steps, and gradually annealed. The result is faster early learning without constraining long-term optimization or incurring excessive VLA inference cost.

We evaluate VLAJS on six challenging manipulation tasks in ManiSkill and validate a subset on a real Franka Panda robot. Focusing on long-horizon objectives and suboptimal reward design, our results demonstrate substantial gains in sample efficiency while producing policies that are directly deployable on real robotic systems.

Therefore, our contributions are:
\begin{enumerate}
    \item We propose a method for accelerating high-frequency, state-based RL using sparse, low-rate guidance from VLA models, producing policies that are directly deployable on real robotic systems \textbf{(C1)}.
    \item We introduce a directional action-consistency regularization that enables flexible, transient guidance without constraining asymptotic performance \textbf{(C2)}.
    \item We demonstrate improved sample efficiency over PPO and distillation-based methods.
    \item We introduce long-horizon and suboptimally rewarded ManiSkill environments for studying RL under difficult credit assignment, which we will publicly release.
\end{enumerate}

\section{Related Work}
\label{sec:related_work}

We review prior work along two axes most relevant to our approach: (i) reinforcement learning under suboptimal credit assignment and expert guidance, and (ii) Vision-Language-Action models for robotic control.

\subsection{RL, Credit Assignment, and Expert Guidance}
\label{subsec:related_work_credit-assignment}

Reinforcement learning is a natural fit for complex manipulation because it supports high-frequency, closed-loop control from state feedback, but it can struggle in long-horizon and sparsely rewarded settings due to inefficient exploration and weak credit assignment~\citep{pignatellisurvey,712192,hil_serl_science,liang2018gpu,schulman2017proximal}.

A common strategy to mitigate poor credit assignment is to incorporate expert guidance~\citep{libardi2021guided,rajeswaran2018learning,maeureka,juelg2025refinedpolicydistillationvla}. Fig.~\ref{fig:sota} summarizes the corresponding formal policy/objective definitions, while Fig.~\ref{fig:guidance} provides an intuitive, trajectory-level view of exploration strategies. Persistent behavioral guidance~\citep{NIPS2013_e034fb6b,ross2011dagger,ross2014dagger2} mixes a learned policy $\pi_\theta$ with an expert policy $\pi_E$,
\[
\pi = (1-\beta)\pi_\theta + \beta\pi_E,
\]
where $\beta>0$ controls expert intervention. Jump-start RL~\citep{uchendu2023jump,5480345,vecerik2017leveraging,hester2018deep} instead applies \emph{transient} behavioral guidance by delegating control to a guide policy $\pi_g$ only early in training,
\[
\pi_t =
\begin{cases}
\pi_g, & t \leq h, \\
\pi_\theta, & t > h,
\end{cases}
\]
where $h$ denotes the number of guide steps.

Complementarily, distillation-style approaches~\citep{juelg2025refinedpolicydistillationvla,xurldg,spigler2024proximal,agarwal2022reincarnating,green2019distillation,czarnecki2019distilling} guide learning via an auxiliary action-matching loss,
\[
\mathcal{L}_{\text{RL}} + \| a_\pi - a_{\text{teacher}} \|^2,
\]
which can overly constrain optimization when applied persistently, and can be brittle when teacher supervision is sparse or imperfect.

These limitations motivate \emph{transient auxiliary guidance}: use a teacher primarily early in training to bias exploration and improve credit assignment, then anneal and remove the auxiliary signal once the on-policy learner begins to improve reliably.

In addition to online guidance, several approaches incorporate prior data or expert policies to improve sample efficiency. Offline-to-online RL pretrains policies or value functions on logged datasets before fine-tuning, while imitation learning/behavior cloning provides initialization but can suffer from distribution shift~\citep{nakamoto2023cal,nair2022learning,reed2022generalist}.

\subsection{Vision-Language-Action Models}
\label{subsec:related_work_vla}

Vision-Language-Action models extend foundation-model paradigms to robotics by directly mapping multimodal observations and language instructions to actions~\citep{firoozi2023foundation,openx2023,brohan2022rt1,brohan2023rt2}. Recent VLAs demonstrate impressive generalization across manipulation tasks, but their practical deployment is constrained by inference latency, low control frequency, and reliance on demonstrations~\citep{zhao2023learning,kim2024openvla,kim2025fine}. Techniques such as action chunking and parallel decoding improve throughput but do not fundamentally address the lack of tight closed-loop feedback~\citep{Black20240AV,intelligence2025pi_,octo2024,liu2024rdt1b}.

Few works explore combining VLAs with downstream RL fine-tuning~\citep{li2025simplevla} or distillation~\citep{xurldg}. Refined Policy Distillation (RPD)~\citep{juelg2025refinedpolicydistillationvla} trains a PPO agent under continuous supervision from a VLA teacher using an action-matching loss. While effective in ideal scenarios, RPD assumes expensive teacher access at every timestep, relies primarily on visual inputs, and produces a standalone policy evaluated only under controlled training conditions, without real-world experiments.

\begin{figure}[t]
    \centering

    % -------- First row: three small + one large --------
    \begin{subfigure}[t]{0.32\linewidth}
        \centering
        \includegraphics[width=\linewidth]{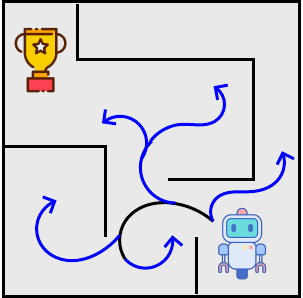}
        \caption{\textbf{Vanilla RL}}
        \label{fig:guidance_a}
    \end{subfigure}
    \hfill
    \begin{subfigure}[t]{0.32\linewidth}
        \centering
        \includegraphics[width=\linewidth]{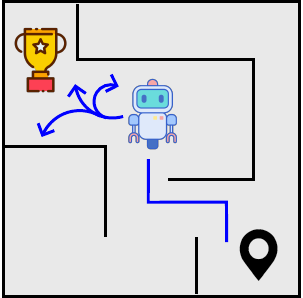}
        \caption{\textbf{Behavioral guidance}}
        \label{fig:guidance_b}
    \end{subfigure}
    \hfill
    \begin{subfigure}[t]{0.32\linewidth}
        \centering
        \includegraphics[width=\linewidth]{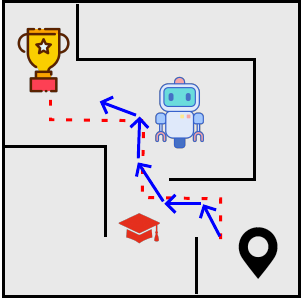}
        \caption{\textbf{Auxiliary guidance}}
        \label{fig:guidance_c}
    \end{subfigure}
    %\hfill
    %\begin{subfigure}[t]{0.5\linewidth}
    %    \centering
    %    \includegraphics[width=\linewidth]{images/Fig-guidance_d.pdf}
    %    \caption{\textbf{Transient auxiliary guidance (VLAJS)}. The auxiliary guidance anneals over time, shaping early exploration while preserving free exploration later on.}
    %    \label{fig:guidance_d}
    %\end{subfigure}

    \caption{
    \textbf{Guidance mechanisms for exploration in RL.} (a) Relies on random exploration. (b) Executes an imitation-learned policy for an initial phase (solid path). (c) Continuously biases learning via a teacher-provided signal (dashed red path) without directly executing actions.
    }
    \label{fig:guidance}
\end{figure}

In contrast, our approach positions the RL agent as a reusable, state-based control layer that is guided—but not dominated—by a VLA. By using sparse, transient auxiliary guidance rather than behavioral (Fig.~\ref{fig:guidance_b}) or persistent (Fig.~\ref{fig:aux-guidance_b}) imitation, VLAJS bridges high-level foundation-model reasoning with efficient, high-frequency control, addressing both credit assignment and real-world execution constraints.

\section{Methodology}
\label{sec:methodology}

In this paper, we propose \textbf{Vision-Language-Action Jump-Starting (VLAJS)}, an on-policy RL method that leverages a pretrained Vision--Language--Action model~\citep{kim2024openvla,octo2024} as \emph{sparse, transient auxiliary guidance}. VLAJS targets settings with \emph{suboptimal credit assignment}---in particular (i) long-horizon tasks and (ii) imperfect reward design---where vanilla on-policy RL often fails to discover rewarding behaviors within practical interaction budgets.

\subsection{Preliminaries: PPO for High-Frequency State Control}
\label{subsec:ppo}
All methods build on Proximal Policy Optimization (PPO) with Generalized Advantage Estimation~\citep{schulman2017proximal}. Let $\pi_\theta(a \mid s)$ denote a stochastic policy and $V_\phi(s)$ a value function. PPO minimizes the clipped surrogate objective
\[
\mathcal{L}_{\mathrm{PPO}}(\theta)=
-\mathbb{E}_t\!\left[
\min\!\left(
r_t(\theta)\hat{A}_t,\;
\mathrm{clip}(r_t(\theta),1-\epsilon,1+\epsilon)\hat{A}_t
\right)\right],
\]
where $r_t(\theta)=\frac{\pi_\theta(a_t\mid s_t)}{\pi_{\theta_{\mathrm{old}}}(a_t\mid s_t)}$ and $\hat{A}_t$ are GAE advantages. PPO is attractive for robotics because it supports stable learning of \emph{high-frequency, closed-loop, state-based control} without demonstrations; however, it is notoriously inefficient when rewards are sparse/delayed or horizons are long~\citep{pignatellisurvey}.

\subsection{Sparse VLA Queries and Temporal Discretization}
\label{subsec:sparse_vla}
We assume access to a pretrained VLA teacher that takes a visual observation and a language instruction and outputs a low-rate delta action $a^{\mathrm{VLA}}$ (translation, rotation, gripper). Querying the teacher at every environment step is impractical for long-horizon rollouts and parallel simulation. Therefore, we query the VLA \emph{sparsely in time}: only a small number of calls per rollout. Each teacher delta is then \emph{temporally discretized} into a short sequence of incremental deltas applied as guidance targets over the next $D$ control steps (linear interpolation for translation and SLERP-style~\citep{Shoemake1985AnimatingRW} interpolation for rotation). Outside these windows, the teacher target is treated as absent and masked out of the auxiliary loss.

This yields guidance that is sparse in two ways: (i) sparse teacher calls across time and (ii) sparse supervision within each rollout, enabling practical wall-clock training while still providing exploration bias early in learning (Fig.~\ref{fig:aux-guidance} explains this concept).

\subsection{Reward-Based Jump-Starting \textbf{(C1)}}
\label{subsec:jumpstart}
\begin{figure}[t]
    \centering

    % ===== COLUMN 1 =====
    \begin{minipage}[t]{0.48\columnwidth}
        \vspace{0pt}
        \centering

        \begin{subfigure}[t]{\linewidth}
            \centering
            \includegraphics[width=\linewidth]{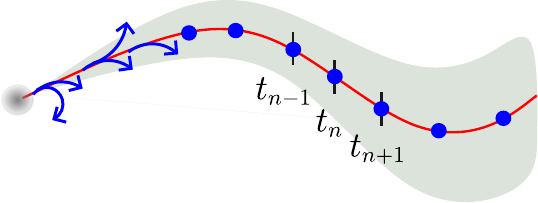}
            \caption{\textbf{Vanilla RL}}
            \label{fig:aux-guidance_a}
        \end{subfigure}

        \vspace{0.5em}

        \begin{subfigure}[t]{\linewidth}
            \centering
            \includegraphics[width=\linewidth]{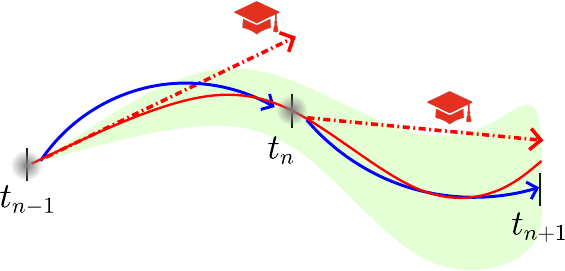}
            \caption{\textbf{Persistent auxiliary}}
            \label{fig:aux-guidance_b}
        \end{subfigure}
    \end{minipage}
    \hfill
    % ===== COLUMN 2 =====
    \begin{minipage}[t]{0.4705\columnwidth}
        \vspace{0pt}
        \centering

        \begin{subfigure}[t]{\linewidth}
            \centering
            \includegraphics[width=\linewidth]{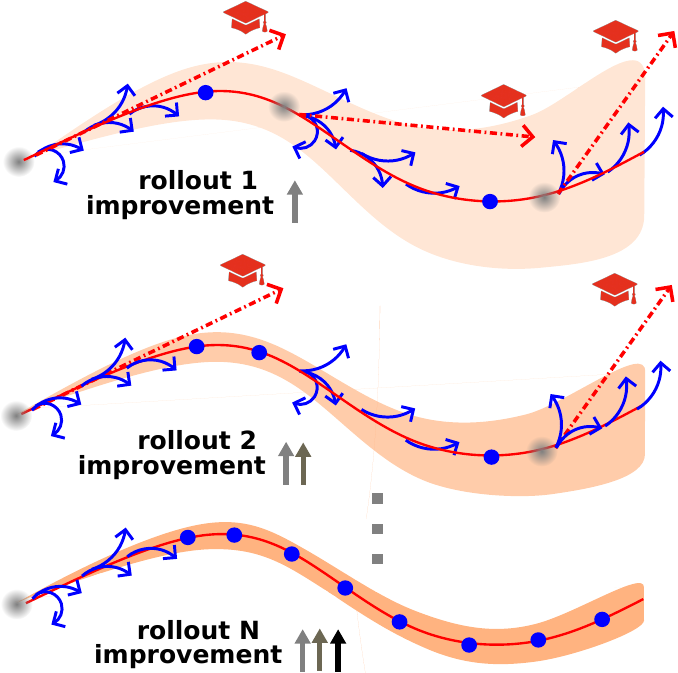}
            \caption{\textbf{Transient auxiliary (VLAJS)}}
            \label{fig:aux-guidance_d}
        \end{subfigure}
    \end{minipage}

    \caption{\textbf{Auxiliary guidance during rollouts.} (a) The policy generates actions solely through on-policy exploration at a fixed control frequency, learning both direction and action scale incrementally from reward. (b) A teacher provides continuous action targets throughout the rollout, constraining both direction and magnitude and forcing the policy to match the teacher’s action scale (distillation/RPD style). (c) Guidance is applied sparsely within a rollout and progressively annealed across rollouts, biasing action direction while allowing the policy to learn its own action magnitude and eventually explore freely.}
    \label{fig:aux-guidance}
\end{figure}

A key principle of VLAJS is that VLA guidance should be \emph{transient}: the teacher is most useful before PPO has discovered a productive exploration regime. Once PPO begins to learn reliably, persistent guidance can become unnecessary (and even harmful if the teacher is suboptimal), and it remains computationally expensive. We therefore introduce a reward-trend--based jump-start mechanism that \emph{reduces} and ultimately \emph{deactivates} teacher usage (see Fig.~\ref{fig:aux-guidance_d}).

\paragraph{Adaptive query rate}
At each PPO iteration, we compute a reward-improvement statistic from a rolling history of mean rollout rewards. As improvement increases, we reduce the number of teacher calls per rollout by an exponential schedule:
\[
N_{\mathrm{calls}} \leftarrow \max(N_{\min}, \lfloor N_{\max}\exp(-\kappa \cdot \Delta \bar{r}) \rfloor),
\]
where $\Delta \bar{r}$ denotes a reward-gain signal computed from recent rollouts (clipped at zero), $\kappa$ controls decay, and $N_{\max}$/$N_{\min}$ bound the calls per rollout. This retains more guidance when PPO is stuck, and quickly sparsifies guidance once learning accelerates.

\paragraph{Permanent deactivation}
We additionally detect \emph{monotonic reward improvement} over a short window of recent iterations and permanently deactivate guidance once the mean rollout reward (``improvement'' in Fig.~\ref{fig:vla-improvement-only}) exceeds a small threshold of $3$. We chose $3$ as the smallest value that reliably marks the onset of meaningful learning: across runs reward often hovers around $\approx 2$ when the policy is still stuck, while once it goes above $3$ learning proceeds reliably, allowing guidance to be turned off as early as possible (see Fig.~\ref{fig:vla-disabled-only}).

\subsection{Directional Action-Consistency Loss \textbf{(C2)}}
\label{subsec:directional_loss}
\begin{figure*}[t]
    \centering
    \begin{subfigure}[t]{0.23\linewidth}
        \centering
        \includegraphics[width=\linewidth]{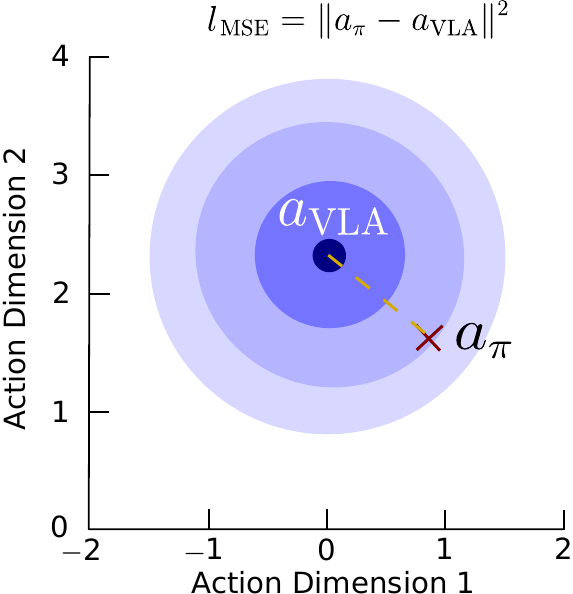}
        \caption{}
        \label{fig:loss_a}
    \end{subfigure}
    \hfill
    \begin{subfigure}[t]{0.23\linewidth}
        \centering
        \includegraphics[width=\linewidth]{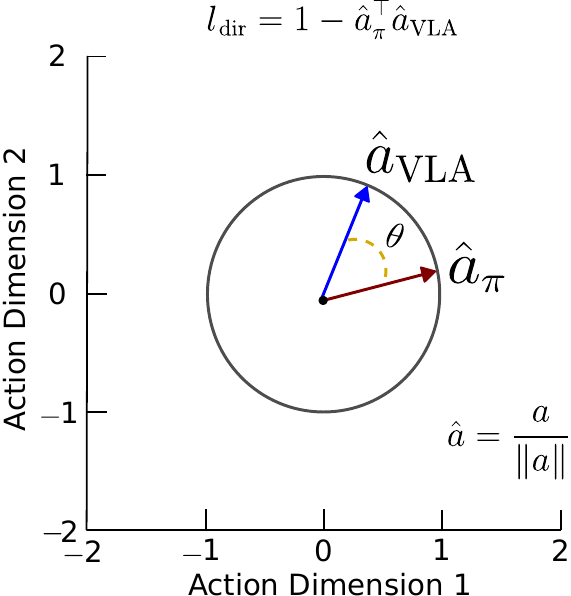}
        \caption{}
        \label{fig:loss_b}
    \end{subfigure}
    \hfill
    \begin{subfigure}[t]{0.23\linewidth}
        \centering
        \includegraphics[width=\linewidth]{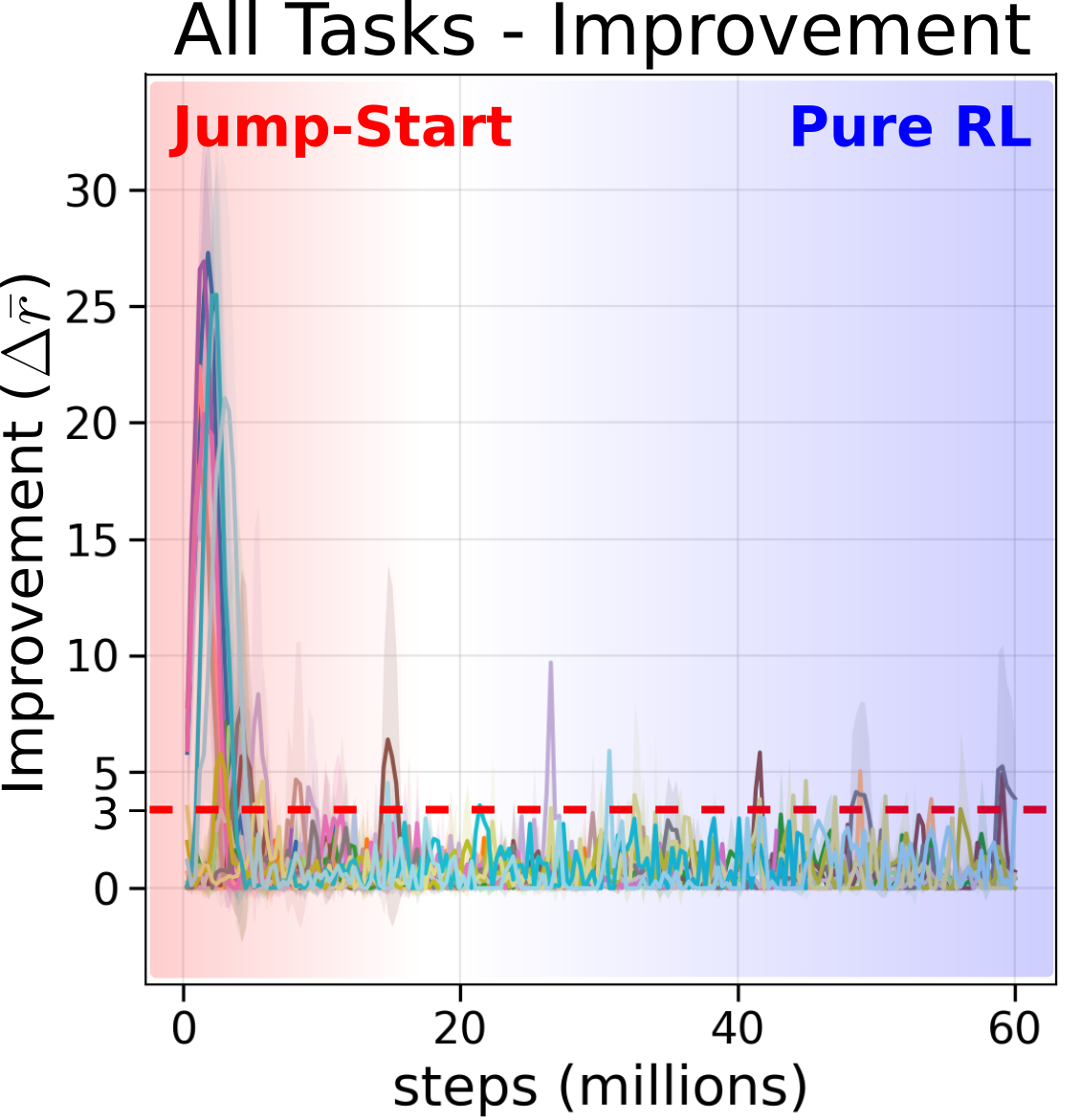}
        \caption{}
        \label{fig:vla-improvement-only}
    \end{subfigure}
    \hfill
    \begin{subfigure}[t]{0.23\linewidth}
        \centering
        \includegraphics[width=\linewidth]{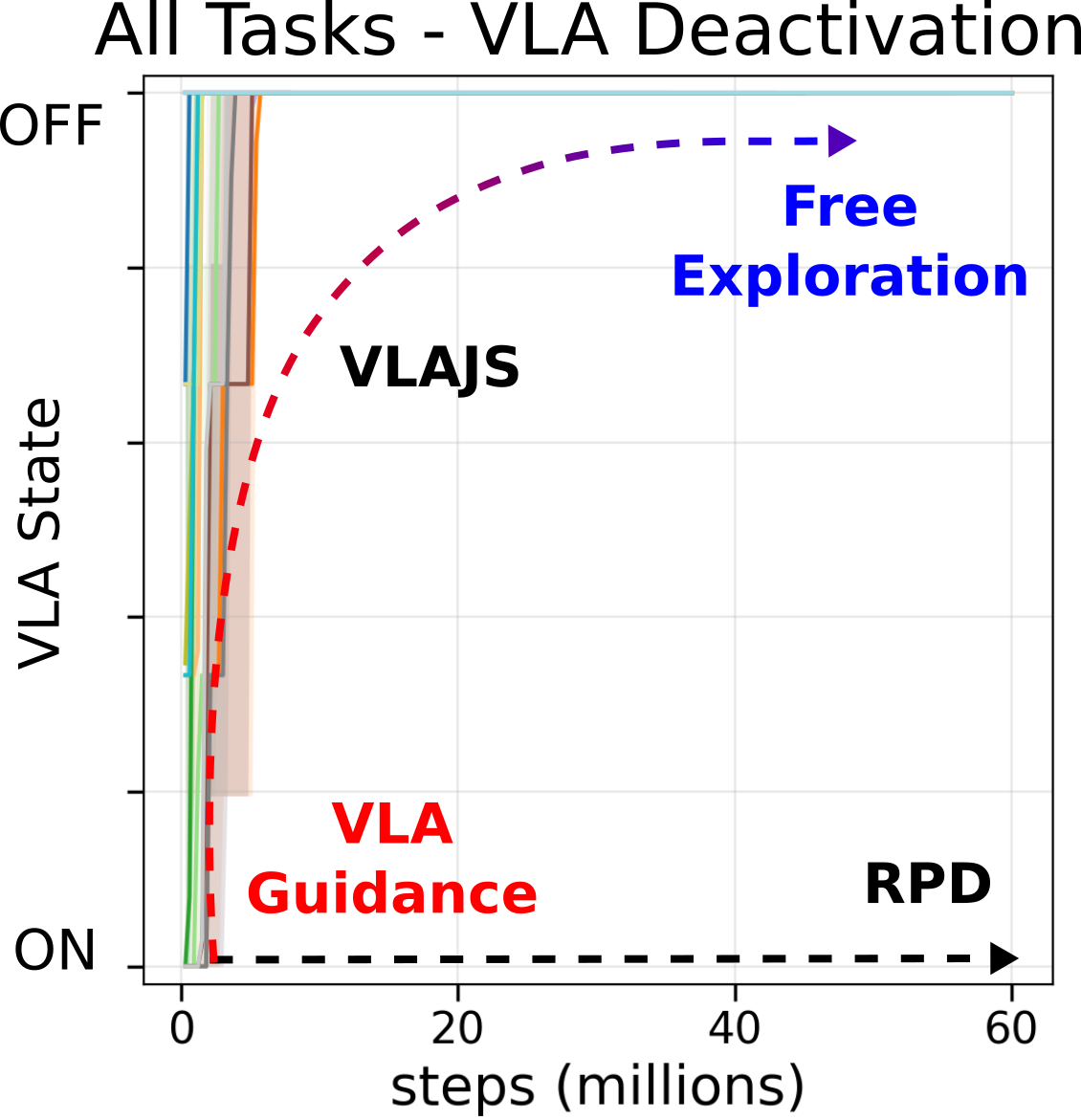}
        \caption{}
        \label{fig:vla-disabled-only}
    \end{subfigure}
    \caption{\textbf{Auxiliary losses for VLA-guided RL.} (a) Distillation-based methods (\textit{e.g.}, RPD) use an MSE loss that penalizes the full Euclidean distance between policy and teacher actions, constraining both action direction and magnitude. (b) VLAJS instead employs a directional action-consistency loss that penalizes angular misalignment between policy and VLA actions, while remaining invariant to action scale. (c–d) Plots show the superimposition of all training runs across tasks, which consistently follow the same trend: VLA guidance produces a pronounced jump-start, visible as a sharp early bump in rollout reward improvement (the metric used for guidance deactivation), after which the signal remains low as guidance is adaptively turned off.}
    \label{fig:loss}
\end{figure*}

Using a VLA teacher as intermittent guidance differs from classical distillation~\citep{juelg2025refinedpolicydistillationvla}: the teacher is queried sparsely and can be \emph{suboptimal} for precise, high-frequency control. In this setting, directly matching teacher actions (\textit{e.g.}, MSE in Fig.~\ref{fig:loss_a}) can be too strong and can inject inconsistent gradients when supervision appears intermittently. Instead, we treat teacher outputs as \emph{directional hints} (see Fig.~\ref{fig:loss_b}).

Let $\mu_\theta(s_t)$ denote the policy mean action and $\tilde{a}^{\mathrm{VLA}}_t$ the discretized teacher target at time $t$ (only present during discretization windows). We split actions into translation and rotation components and define a cosine misalignment loss
\[
\ell_{\mathrm{dir}}(x,y)=1-\frac{\langle x,y\rangle}{\|x\|\|y\|+\varepsilon}.
\]
Our auxiliary objective is
%\[
%\mathcal{L}_{\mathrm{dir}}
%=
%\mathbb{E}_t\!\left[
%\mathbf{1}[\mathrm{valid}_t]\left(
%\ell_{\mathrm{dir}}(\mu^{\mathrm{pos}}_\theta(s_t), \tilde{a}^{\mathrm{pos}}_t)
%+
%\ell_{\mathrm{dir}}(\mu^{\mathrm{rot}}_\theta(s_t), \tilde{a}^{\mathrm{rot}}_t)
%\right)\right],
%\]
\[
\mathcal{L}_{\mathrm{dir}}=
\mathbb{E}_t\!\left[\mathbf{1}[\mathrm{valid}_t]\sum_{c\in\{\mathrm{pos},\mathrm{rot}\}}\ell_{\mathrm{dir}}\big(\mu^{c}_\theta(s_t), \tilde{a}^{c}_t\big)\right],
\]
where $\mathbf{1}[\mathrm{valid}_t]$ masks timesteps without guidance and we skip components with near-zero teacher vectors to avoid unstable normalization. We do not constrain the gripper dimension in the auxiliary loss.

\paragraph*{Rationale for direction-only}
Cosine alignment preserves the \emph{direction} suggested by the teacher while allowing PPO to choose action magnitudes and fine corrections. This is particularly important when (i) teacher actions are discretized across multiple steps, (ii) teacher scale may not match the student's control frequency, and (iii) the teacher is imperfect and should not be copied exactly.

\subsection{Training Objective}
\label{subsec:objective}
We augment PPO updates with the auxiliary guidance loss:
\[
\mathcal{L}(\theta)=\mathcal{L}_{\mathrm{PPO}}(\theta) + \lambda_t \mathcal{L}_{\mathrm{aux}}(\theta),
\]
where $\mathcal{L}_{\mathrm{aux}}=\mathcal{L}_{\mathrm{dir}}$ for VLAJS. The coefficient $\lambda_t$ follows the same reward-trend schedule used for guidance sparsification and is set to zero after deactivation.

\subsection{Baselines Implemented in Our Code}
\label{subsec:baselines_methods}
We implement three algorithms:

\paragraph{PPO}
Standard PPO~\citep{schulman2017proximal} trained from scratch using only environment reward.

\paragraph{Sparse RPD}
For long-horizon experiments, we additionally evaluate a \emph{persistent} sparse-guidance baseline that queries the teacher sparsely \emph{throughout training} (no deactivation), reflecting ``sparse distillation'' as a computationally feasible alternative to full RPD~\citep{juelg2025refinedpolicydistillationvla} when horizons are long.

\paragraph{VLAJS (RPD)}
An ablation that keeps \emph{exactly the same} sparse query mechanism and jump-start deactivation, but replaces directional guidance with an RPD-style MSE action-matching loss on guided steps:
\[
\mathcal{L}_{\mathrm{MSE}}=
\mathbb{E}_t\big[\mathbf{1}[\mathrm{valid}_t]\ \|\mu_\theta(s_t)-\tilde{a}^{\mathrm{VLA}}_t\|_2^2\big].
\]
This isolates the effect of the directional loss under sparse, transient teacher usage.

\begin{algorithm}[t]
\caption{Vision-Language-Action Jump-Starting}
\label{alg:vlajs}
\KwIn{PPO policy $\pi_\theta$, value $V_\phi$, VLA teacher $\pi_{\mathrm{VLA}}$, rollout horizon $H$, discretization length $D$}
Initialize reward history buffer; \texttt{vla\_disabled}$\leftarrow$False\;
\For{each PPO iteration $k$}{
  Compute reward-gain $\Delta \bar{r}$ from history\;
  Set $N_{\mathrm{calls}}$ and $\lambda_k$ via exponential decay\;
  \If{\texttt{vla\_disabled}}{set $N_{\mathrm{calls}}\!=\!0$, $\lambda_k\!=\!0$;}
  Collect rollout $\{(s_t,a_t,r_t)\}_{t=1}^H$ with PPO\;
  Query VLA at $N_{\mathrm{calls}}$ timesteps; discretize each teacher delta into $D$ targets $\tilde{a}^{\mathrm{VLA}}_t$\;
  Compute GAE advantages and PPO losses\;
  Update $(\theta,\phi)$ using $\mathcal{L}_{\mathrm{PPO}}+\lambda_k \mathcal{L}_{\mathrm{aux}}$\;
  Update reward history; if recent rewards are monotonically improving and $\Delta \bar{r}>3$, set \texttt{vla\_disabled}$\leftarrow$True\;
}
\end{algorithm}

\section{Experimental Evaluation}
\label{sec:methodology_experiments}
\begin{figure*}[t]
    \centering
    \includegraphics[width=\linewidth]{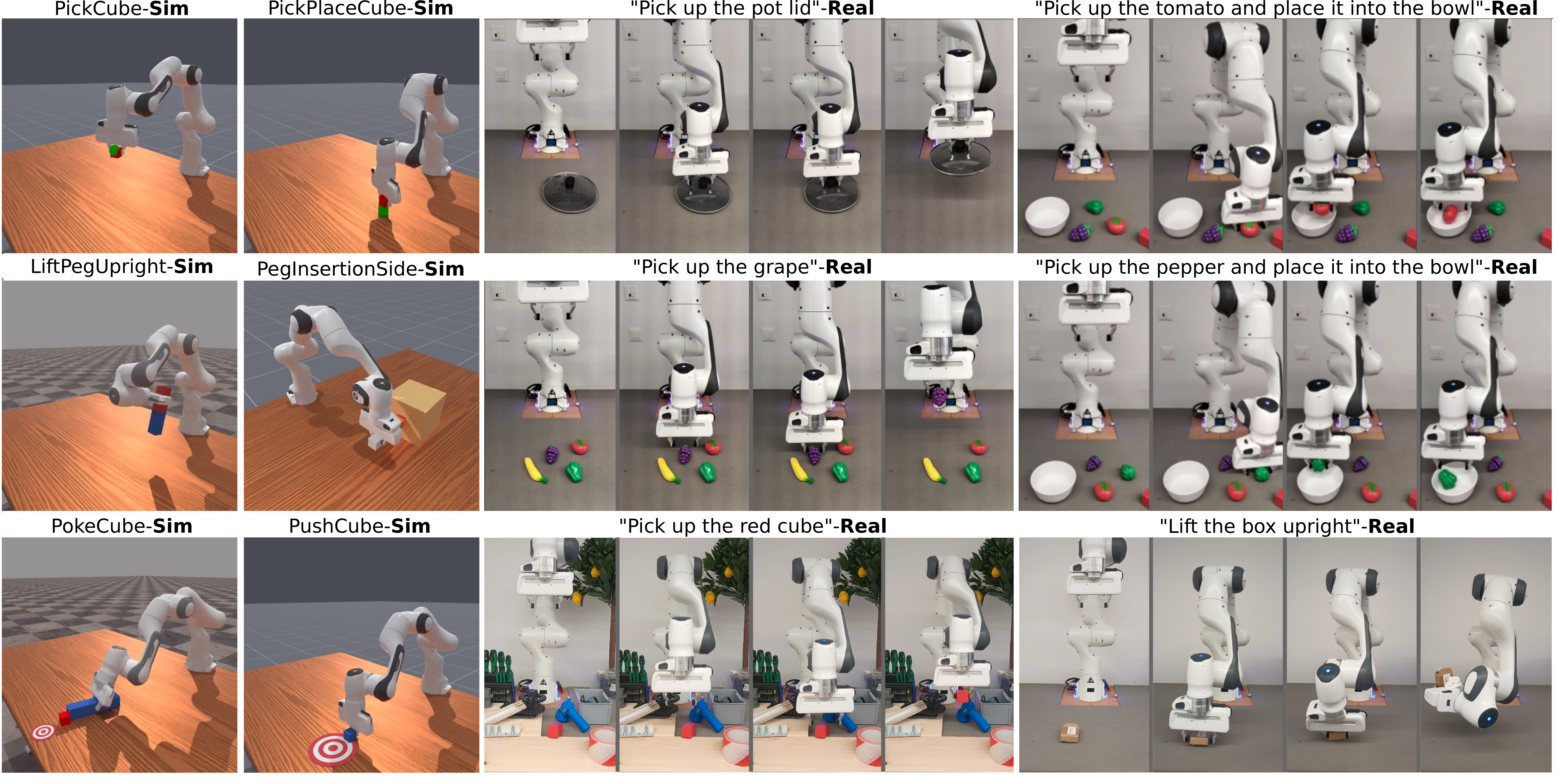}
    \caption{Simulation and real-world manipulation tasks used in our evaluation. 
    Left: six ManiSkill simulation tasks (PickCube, PickPlaceCube, LiftPegUpright, PegInsertionSide, PokeCube, PushCube). 
    Right: zero-shot real-world deployment on a Franka Panda robot across diverse language-specified tasks.}
    \label{fig:environments}
\end{figure*}

We evaluate our approach under two complementary sources of suboptimal credit assignment:

\paragraph{Use Case 1 - Long-horizon task}
We test whether \emph{sparse VLA guidance} is a practical and effective substitute for dense distillation when episodes become very long (\textit{e.g.}, high-frequency control or extended horizons). Here, the goal is primarily to quantify exploration benefits and computational feasibility. We compare \textbf{Sparse RPD}---the building block of VLAJS (RPD)---to PPO.

\paragraph{Use Case 2 - Suboptimal reward design}
We test whether \emph{jump-starting with transient guidance} and a \emph{directional loss} improves learning when rewards are sparse or imperfectly shaped, reflecting realistic reward design constraints. Here we compare \textbf{PPO}, \textbf{VLAJS (RPD)}, and \textbf{VLAJS}.

\subsection{Primary Metrics}
\label{subsec:metrics}
We report per-task success-based metrics aligned with the summary tables:

\paragraph{Success Rate at $t^\ast$ (SR$_{t^\ast}$)}
SR$_{t^\ast}$ measures the fraction of evaluation episodes that successfully complete within a task-specific interaction budget $t^\ast$ (reported in environment steps). This captures whether a method learns the task within a practical sample budget.

\paragraph{Area Under the Success Curve (AUC)}
AUC integrates success rate over the full training budget $[0,B]$ (where $B$ is the total environment-step budget of the experiment). AUC captures both \emph{learning speed} and final performance.

We report bootstrap 95\% confidence intervals across random seeds where available, and macro-averages across tasks.

\subsection{Summary Tables}
\label{subsec:tables}
For long-horizon experiments, Tab.~\ref{tab:lll_success_small_merged} reports a macro-average summary of SR$_{t^\ast}$ and AUC across four extended-horizon tasks, as a preliminary investigation of Sparse RPD. For suboptimal reward experiments, Tab.~\ref{tab:success_tstar_auc_one_row} reports the same metrics for PPO, VLAJS (RPD), and VLAJS across tasks and a macro-average, summarizing Fig.~\ref{fig:res-r}.

\section{Training Setup}
\label{sec:exp_setup}

\subsection{Simulation Environments, Observations, and Actions}
\label{subsec:env_setup}
All simulations are conducted in ManiSkill manipulation environments~\citep{taomaniskill3,xiang2020sapien} (see Fig.~\ref{fig:environments}). The RL policy is a state-based controller: observations include robot proprioception and privileged simulator state (\textit{e.g.}, object poses). Actions are continuous delta end-effector controls (translation and rotation) with a gripper command, executed at a high control frequency. This setting reflects the regime where RL excels at precise closed-loop control but struggles with exploration and long-horizon credit assignment.

\subsection{Teacher Models and Sparse Querying}
\label{subsec:teacher_setup}
We use a pretrained VLA---\textit{OpenVLA-best} with average s.r. of 40\%---as an external teacher that maps RGB observations and a language instruction to a delta action. Due to inference cost, the teacher is queried only a few times per rollout (max. 20\% of the total). Each teacher delta is discretized into a short sequence of incremental deltas over $D$ steps, producing sparse guidance targets within the rollout and zero targets elsewhere. Teacher actions are never executed directly in the environment; they are used only in auxiliary losses during training.

\subsection{Use Case 1 - Long-Horizon Protocol}
\label{subsec:long_horizon_setup}
To isolate the impact of long horizons, we take standard ManiSkill tasks and increase the effective horizon length by $10\times$. This models realistic scenarios where policies operate at higher frequencies or where tasks require extended action sequences, amplifying the difficulty of exploration and reward propagation.

We focus on the feasibility and benefit of \emph{persistent sparse guidance} and therefore compare:
(i) PPO and (ii) Sparse RPD variants (teacher queried sparsely throughout training). We report SR$_{t^\ast}$ and AUC under fixed step budgets $B$ (Tab.~\ref{tab:lll_success_small_merged}).

\subsection{Use Case 2 - Suboptimal Reward Design Protocol}
\label{subsec:subreward_setup}
To model realistic reward engineering constraints, we modify ManiSkill reward functions into simplified, more intuitive variants that provide weaker shaping (\textit{e.g.}, sparse success signals such as rewarding only object pickup rather than dense shaping). This induces suboptimal credit assignment even for tasks that are otherwise solvable with dense rewards.

In this use case, we evaluate whether \emph{transient jump-start guidance} and \emph{directional regularization} improve sample efficiency. We compare:
(i) PPO, (ii) VLAJS (RPD), and (iii) VLAJS (ours). Performance is summarized via SR$_{t^\ast}$ and AUC with task-specific $t^\ast$ and macro-averages (Tab.~\ref{tab:success_tstar_auc_one_row}).

\subsection{Training Details and Reproducibility}
\label{subsec:training_details}
All methods share the same PPO backbone, network architecture, optimizer settings, and rollout configuration. Guidance-related hyperparameters (maximum calls per rollout, decay rate, discretization length $D$, and deactivation patience) are fixed across tasks. We evaluate each method with multiple random seeds and report mean performance with bootstrap confidence intervals where applicable.

\section{Results}
\label{sec:results}

We evaluate our approach under two complementary sources of suboptimal credit assignment: \textbf{long-horizon tasks} and \textbf{imperfect reward design}. These settings isolate different failure modes of on-policy RL and motivate different comparisons.

\subsection{Use Case 1 - Long-Horizon Task}
\label{subsec:results_long_horizon}
\begin{figure}[t]
    \centering
    \includegraphics[width=\linewidth]{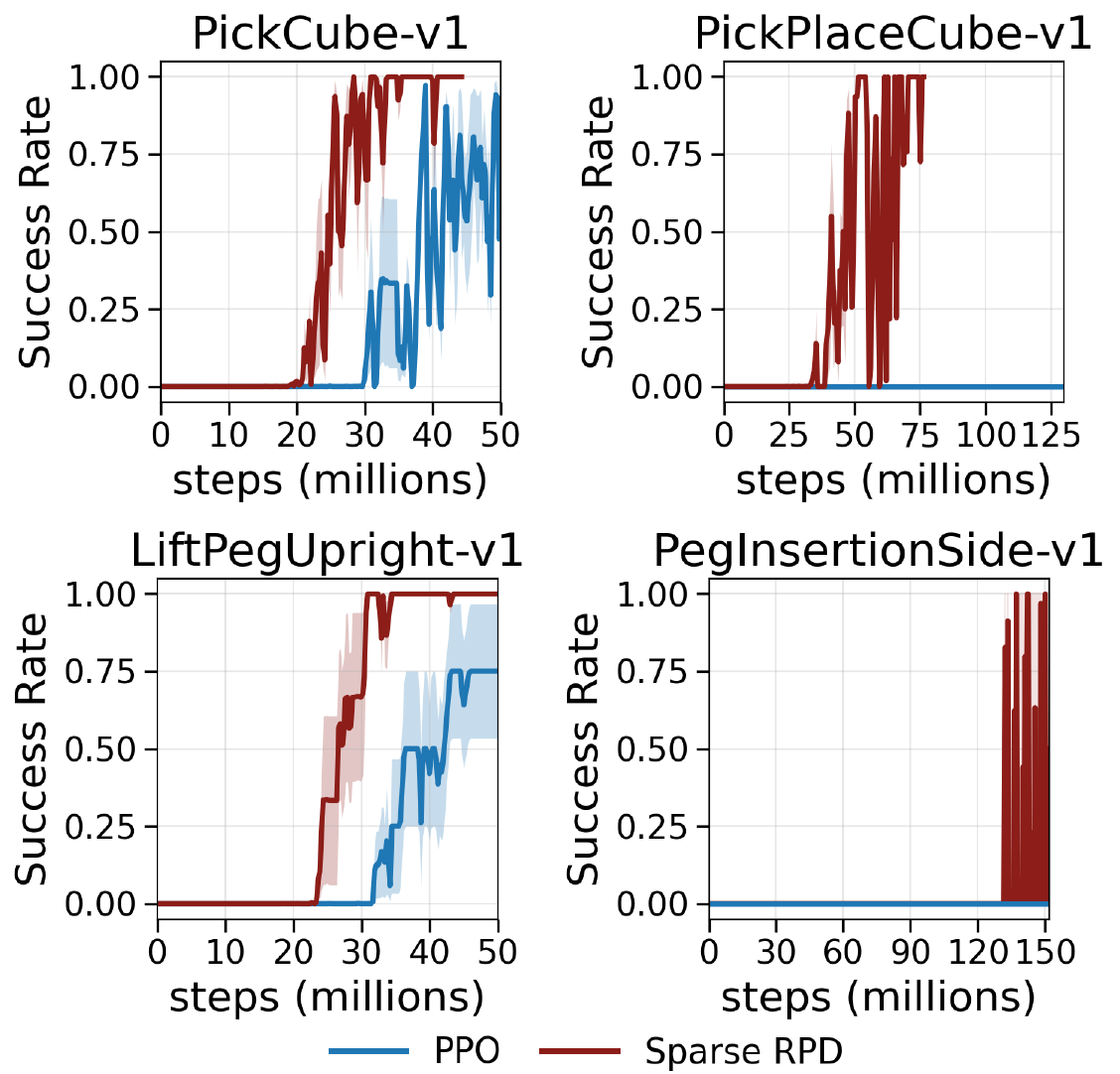}
    \caption{Learning curves for long-horizon tasks. Sparse RPD makes distillation feasible in long-horizon tasks significantly accelerating convergence compared to PPO baselines. Standard RPD is omitted due to prohibitive training time.}
    \label{fig:res-lh}
\end{figure}

\label{subsec:results_suboptimal_rewards}
\begin{figure*}[t]
    \centering
    \includegraphics[width=\linewidth]{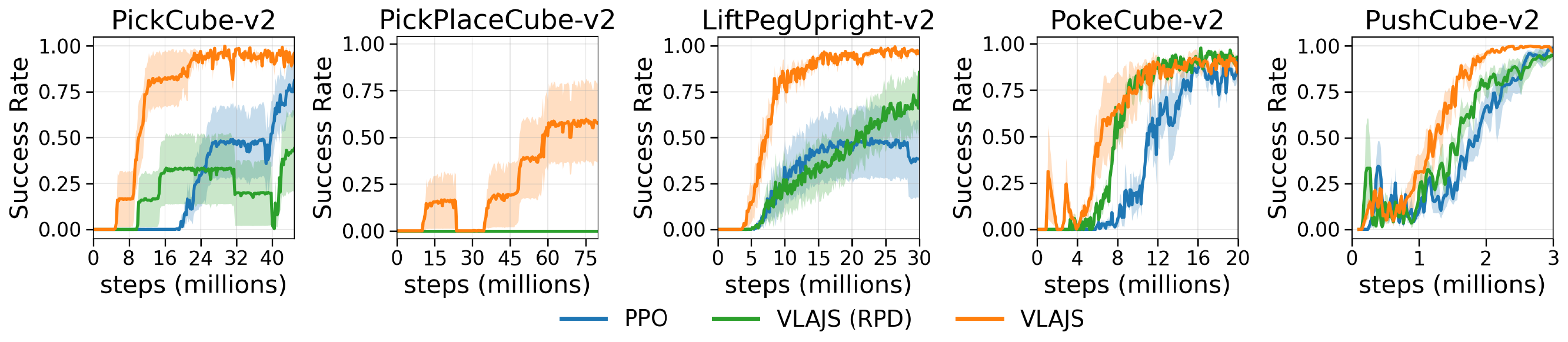}
    \caption{Learning curves and sample-efficiency comparison for suboptimal reward tasks. VLAJS consistently outperforms PPO and distillation-based baselines---VLAJS (RPD). The initial increase in return is driven by the VLA-based jump-start. Once the VLA guidance is deactivated, the agent transitions to a purely learned policy, which results in a temporary reduction in return while maintaining or improving task success (see the additional material for return plots).}
    \label{fig:res-r}
\end{figure*}

\input{tables/sub_reward_tables}

We first study environments with extended episode horizons, which amplify delayed reward propagation and make exploration particularly challenging for PPO. The objective of this experiment is to assess whether \emph{sparse} VLA guidance is computationally feasible and beneficial when dense teacher supervision is impractical (RPD~\citep{juelg2025refinedpolicydistillationvla}).

Tab.~\ref{tab:lll_success_small_merged} reports macro-averaged success rate at a fixed interaction budget ($\mathrm{SR}_{t^\ast}$) and area under the success curve (AUC) across four long-horizon tasks. Across all tasks, Sparse RPD consistently outperforms PPO, often by a large margin in both early success and AUC (see Fig.~\ref{fig:res-lh}).

These results demonstrate that even very sparse auxiliary guidance from a VLA provides strong exploration benefits in long-horizon regimes, while remaining computationally tractable. This establishes sparse guidance as a viable building block, but does not yet address whether the agent can learn beyond the teacher or whether persistent supervision is desirable.

\subsection{Use Case 2 - Suboptimal Reward Design}
We next consider environments with deliberately simplified and sparse reward definitions, designed to reflect realistic reward engineering constraints. In these settings, PPO often fails to learn meaningful behaviors within practical interaction budgets, despite moderate episode horizons.

Tab.~\ref{tab:success_tstar_auc_one_row} compares PPO, VLAJS (RPD), and VLAJS using success rate at $t^\ast$ and AUC. Unlike the long-horizon case with sparse persistent guidance, imitation-based loss is no longer sufficient when jump-starting: VLAJS (RPD) provides limited or inconsistent improvements over PPO, and in several tasks fails to meaningfully accelerate learning.

In contrast, VLAJS consistently achieves higher success rates (results in Fig.~\ref{fig:res-r}) and larger AUC across all tasks, including an out-of-distribution (OOD) task that the VLA was not trained on (\textit{PushCube-v2}). By combining sparse guidance with reward-aware deactivation and a directional action-consistency loss, VLAJS effectively jump-starts learning while preserving the ability to optimize beyond the teacher. Notably, performance often continues to improve after VLA guidance is fully deactivated, indicating that the learned policy is not constrained by the teacher.

\subsection{Zero-shot Real-World Deployment}
\begin{figure}[b!]
    \centering
    \includegraphics[width=0.8\columnwidth]{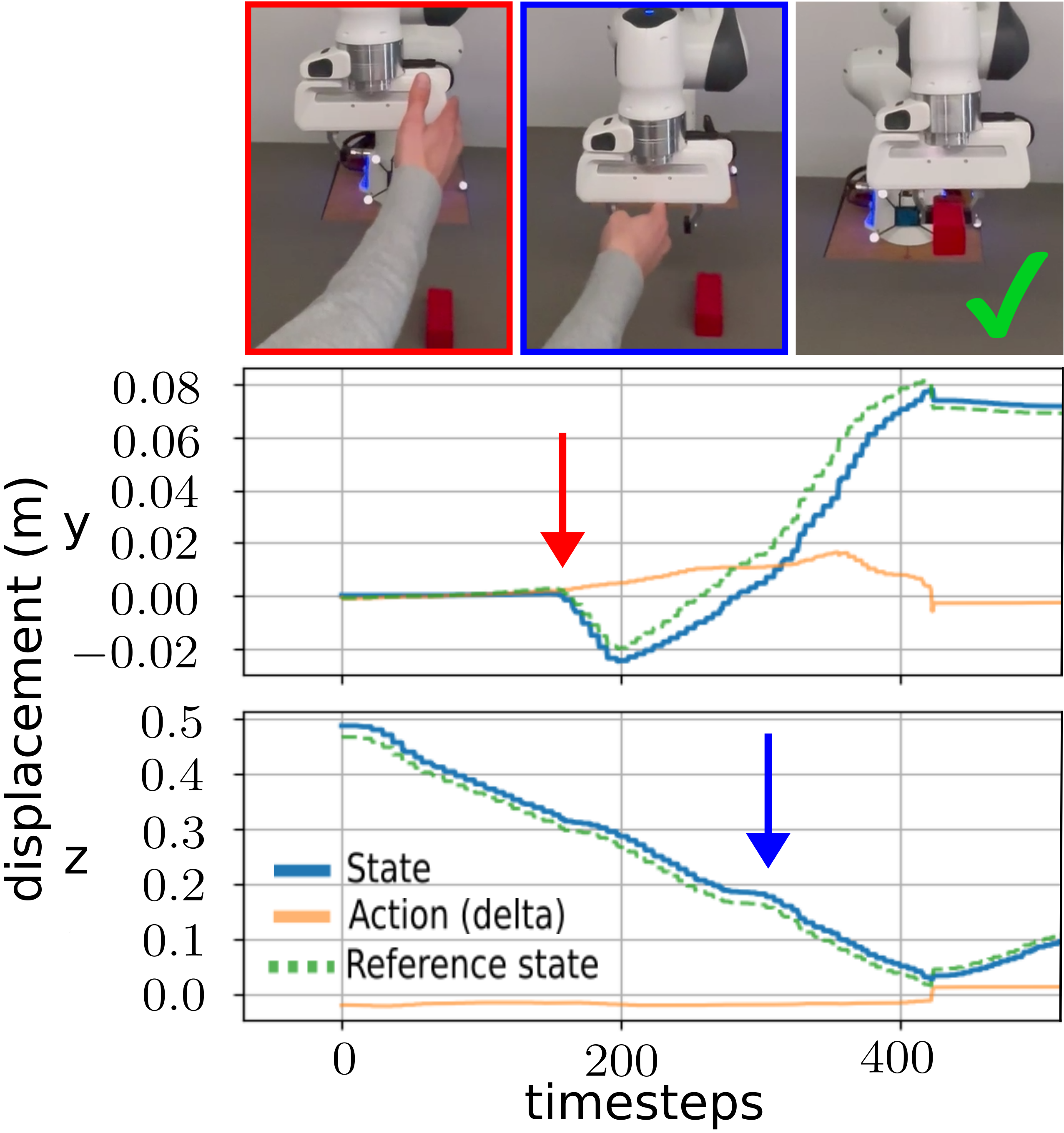}
    \caption{Policy robustness under external perturbations and clutter. VLAJS maintains performance where VLA-only policies fail.}
    \label{fig:simtoreal-perturbation}
\end{figure}

Real-world results for randomly sampled objects (see supplementary material for the full list) from a fixed subset of tasks are summarized in Table~\ref{tab:real_robot}, demonstrating \emph{zero-shot} deployment of the learned policies. Figure~\ref{fig:environments} shows performance under randomized visual conditions, highlighting robustness enabled by the state-based policy and a visually robust detector. In contrast to a VLA baseline, which fails under strong visual perturbations (\textit{e.g.}, a human hand entering the scene), our policies remain stable and successfully complete the task, as shown in Figure~\ref{fig:simtoreal-perturbation}. Real-world state estimates are obtained using a pretrained YOLO detector~\cite{redmon2016you}; additional details are provided in the supplementary material.

\input{tables/real_world.tex}

\section{Discussion}
\label{sec:discussion}

\input{tables/long_horizon_small_table}

\begin{figure}[t!]
    \centering
    \begin{subfigure}[t]{0.48\columnwidth}
        \centering
        \includegraphics[width=\linewidth]{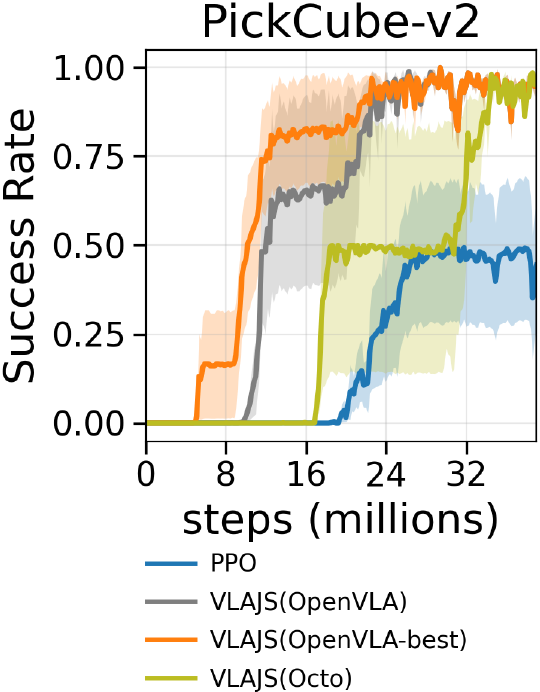}
        \caption{Comparing different VLA teachers and different levels of their fine-tuning (OpenVLA - 10\% s.r., OpenVLA-best - 40\% s.r. and Octo - 10\% s.r.) when used within VLAJS. Success rates averaged across tasks.}
        \label{fig:vla_comparison}
    \end{subfigure}
    \hfill
    \begin{subfigure}[t]{0.48\columnwidth}
        \centering
        \includegraphics[width=\linewidth]{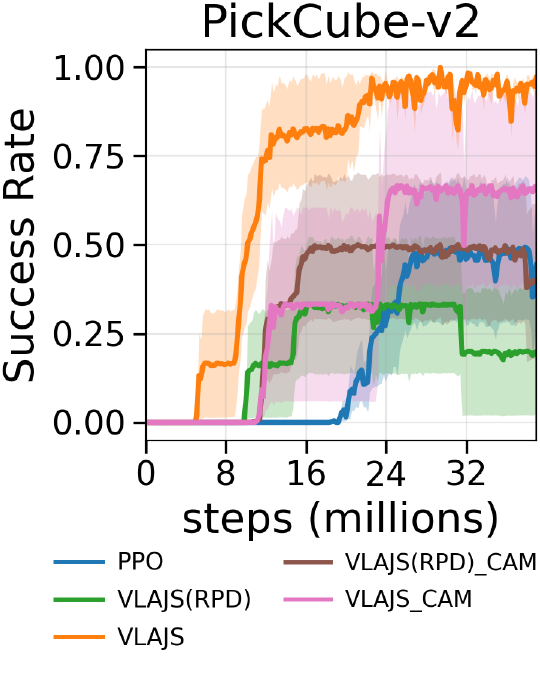}
        \caption{Comparing different VLA camera views for OpenVLA. \textit{VLAJS(RPD)\_CAM} and \textit{VLAJS\_CAM} indicate the performance of our approach when the VLA camera is in a new position (OOD), versus the baselines.}
        \label{fig:vla_cam_comparison}
    \end{subfigure}

    \caption{Comparisons on VLA teachers.}
    \label{fig:vla-comparison}
\end{figure}

Our results show that \emph{when} and \emph{how} teacher signals are used matters as much as the teacher itself, revealing complementary roles of VLA guidance in on-policy RL and enabling zero-shot real-world deployment on a real robot (Tab.~\ref{tab:real_robot}). In long-horizon tasks, where delayed reward propagation hinders exploration, sparse but persistent auxiliary guidance provides a practical and deployable alternative to dense distillation, yielding substantial gains over PPO (Tab.~\ref{tab:lll_success_small_merged}). We then consider a harder setting with suboptimal reward definitions to directly test whether auxiliary guidance can be made \emph{transient}, reducing reliance on teacher queries and overall computation. In this regime, we find that while persistent guidance may still be effective, distillation-style action matching--VLAJS (RPD)--is no longer suitable for jump-starting learning, motivating the use of a \emph{weak, directional} consistency loss that bootstraps task-relevant exploration without over-constraining the policy (Tab.~\ref{tab:success_tstar_auc_one_row}). As a result, policies trained with VLAJS continue to improve after guidance is fully deactivated (Fig.~\ref{fig:res-r}). Finally, our framework is compatible with arbitrary VLA teachers. Due to the high cost of VLA fine-tuning, we evaluate two representative models---OpenVLA and Octo, a diffusion policy VLA. We show that even weak or OOD teachers can accelerate learning, surprisingly suggesting that VLA performance is not critically important in VLAJS (Fig.~\ref{fig:vla_comparison}). The framework also remains robust to changes in the observation setup (Fig.~\ref{fig:vla_cam_comparison}).

\section{Limitations}
\label{sec:limitations}

While VLAJS improves sample efficiency in difficult credit-assignment regimes, it still relies on a VLA teacher that provides at least minimally reliable directional cues. Although VLAJS is relatively insensitive to the teacher’s success rate (see \ref{sec:discussion}), in practice current VLAs often require environment-specific fine-tuning to be useful, and obtaining such adaptation can be expensive.

Using a large VLA during training also introduces nontrivial wall-clock overhead and systems complexity, including GPU memory pressure, inference latency, and engineering effort for external serving. The practical benefit therefore depends on the trade-off between fewer environment interactions and higher per-step compute.

Finally, our experiments focus on tabletop manipulation with privileged simulator state for the RL policy and RGB observations for the teacher. Extending VLAJS to fully vision-based RL, force-interactive manipulation, or longer-horizon multi-stage tasks may require additional components (\textit{e.g.}, hierarchy, memory, or state estimation). We also rely on a simple reward-based heuristic to deactivate guidance, which may be brittle in highly stochastic settings; more principled uncertainty- or advantage-based gating could improve robustness.

\section{Conclusion}
\label{sec:conclusion}

We presented \textbf{VLAJS}, which improves the sample efficiency of on-policy reinforcement learning for robotic manipulation by leveraging pretrained VLA models as \emph{sparse, transient auxiliary guidance}. VLAJS combines (i) a \textbf{reward-based jump-start schedule} that reduces and permanently deactivates teacher usage, and (ii) a \textbf{directional action-consistency loss} that interprets VLA outputs as coarse directional hints rather than strict action targets. 

Across long-horizon tasks and suboptimal reward designs, our approach accelerates learning and improves final performance relative to PPO and RPD-based sparse distillation baselines, while enabling continued policy improvement after guidance is removed. To the best of our knowledge, we are the first to deploy \emph{jump-started policies guided by a VLA} on a real robotic system, demonstrating zero-shot transfer to a Franka Panda robot.

Future work will focus on reducing teacher overhead by querying VLA guidance only when needed, extending the approach to vision-based RL for more complex manipulation and navigation tasks, and exploring direct real-world fine-tuning of RL policies with the VLA model.

\section*{Acknowledgments}
The research leading to these results was supported by the Swiss Drone and Robotics Centre of the Department of Defence, Civil Protection and Sport, armasuisse S+T under project n°050-44. This work was also supported by the Swiss AI Initiative and utilized computing resources from the Swiss National Supercomputing Centre (CSCS) on the Alps system under project a144.

%% Use plainnat to work nicely with natbib. 

\bibliographystyle{plainnat}
\bibliography{references}

\end{document}

%% file: tables/sub_reward_tables.tex
% Requires: \usepackage{booktabs}
\begin{table*}[t]
\centering
\footnotesize
\setlength{\tabcolsep}{4.5pt}
\renewcommand{\arraystretch}{0.95}

\begin{tabular}{lcccc}
\toprule
Task & $t^\ast$ &
PPO (SR\_$t^\ast$/AUC $\uparrow$) $(\%)$ &
VLAJS (RPD) (SR\_$t^\ast$/AUC $\uparrow$) $(\%)$ &
\textbf{VLAJS} (SR\_$t^\ast$/AUC $\uparrow$) $(\%)$ \\
\midrule

PickCube-v1      & 2.5M  &
88.7 [83.3,94.4] / 72.4 [71.5,73.8] &
59.0 [33.3,76.9] / 71.0 [68.7,73.5] &
\textbf{91.7 [90.0,93.3]} / \textbf{72.9 [72.2,74.0]} \\

PickCube-v2      & 9.9M  &
0.0 / 59.6 [52.8,69.2] &
1.1 [0.0,2.9] / 76.7 [56.1,89.3] &
\textbf{95.1 [92.3,100.0]} / \textbf{88.7 [87.4,90.2]} \\

PickPlaceCube-v1 & 11.4M &
45.4 [0.0,93.3] / 70.3 [67.5,74.0] &
\textbf{86.7 [60.0,100.0]} / \textbf{72.8 [70.2,76.7]} &
39.1 [0.0,93.8] / 72.3 [65.1,81.3] \\

PickPlaceCube-v2 & 37.7M &
0.0 / 0.0 &
0.0 / 0.0 &
\textbf{65.9 [0.0,100.0]} / \textbf{66.0 [50.7,86.2]} \\

LiftPegUpright-v1 & 2.0M &
\textbf{84.8 [63.6,100.0]} / 85.9 [84.1,87.3] &
0.8 [0.0,1.3] / 81.1 [76.7,85.1] &
80.3 [50.0,100.0] / \textbf{87.3 [86.3,88.4]} \\

LiftPegUpright-v2 & 7.3M  &
16.9 [0.0,36.4] / 76.6 [70.4,80.5] &
0.3 [0.0,1.0] / 45.3 [17.8,60.0] &
\textbf{91.5 [85.7,100.0]} / \textbf{80.5 [76.6,83.0]} \\

LiftPegUpright-v3 & 17.1M &
13.2 [0.0,20.0] / 58.4 [39.8,67.9] &
19.2 [0.0,37.5] / 33.6 [0.0,59.7] &
\textbf{63.3 [0.0,100.0]} / \textbf{79.1 [66.0,90.3]} \\

PokeCube-v2      & 8.8M  &
9.9 [0.0,16.7] / 36.9 [25.6,43.3] &
75.1 [64.8,92.3] / 54.8 [52.7,57.5] &
\textbf{81.4 [50.0,100.0]} / \textbf{60.7 [43.8,70.2]} \\

PushCube-v2 (OOD) & 1.9M  &
49.1 [26.7,61.1] / 96.5 [95.7,96.9] &
75.6 [65.2,87.8] / 97.7 [97.5,97.9] &
\textbf{94.4 [92.7,96.7]} / \textbf{98.4 [98.2,98.6]} \\

\midrule
Macro Avg & -- &
34.2 / 61.8 &
35.3 / 59.3 &
\textbf{78.1} / \textbf{78.4} \\
\bottomrule
\end{tabular}

\caption{Per-task performance reported as \textbf{Success Rate} at \textbf{$t^\ast$ (SR\_$t^\ast$)} / \textbf{Area Under the Success Curve (AUC$_{0..B}$)}, both in percent.
SR\_$t^\ast$ measures the fraction of evaluation episodes that successfully complete the task within the task-specific time budget $t^\ast$ (in environment steps).
AUC$_{0..B}$ integrates the success rate over the full episode horizon $[0,B]$, capturing both final performance and learning speed.
Bracketed values denote bootstrap 95\% confidence intervals over random seeds ($n=6$). Bold numbers indicate the best mean performance among the baselines.}
\label{tab:success_tstar_auc_one_row}
\end{table*}

%% file: tables/real_world.tex
\begin{table}[t]
\centering
\begin{tabular}{lccc}
\toprule
Policy & Lift Cube & Pick \& Place & Peg Reorientation \\
\midrule
OpenVLA-best         & 47\% & 40\% & -- \\
VLAJS (zero-shot) & 70\% & 80\% & 20\% \\
\bottomrule
\end{tabular}
\caption{Real robot deployment success rates (20 trials per task).}
\label{tab:real_robot}
\end{table}

%% file: tables/long_horizon_small_table.tex
%%%%%%%%%%%%%%%%%%%

\begin{table}[t]
\centering
\scriptsize
\setlength{\tabcolsep}{2.2pt}
\renewcommand{\arraystretch}{0.92}
\begin{tabular}{lcccc}
\toprule
Algorithm & SR\_$t^\ast$ $\uparrow$ $(\%)$ & Wins & AUC $\uparrow$ $(\%)$ & Wins \\
\midrule
PPO & 0.0 & 0 & 7.2 & 0 \\
Sparse RPD (OpenVLA) & 8.3 & 1 & 12.8 & 0 \\
Sparse RPD (OpenVLA-best) & \textbf{40.3} & 3 & \textbf{37.6} & 4 \\
\bottomrule
\end{tabular}
\caption{Macro averages across the 4 long-horizon tasks. Sparse RPD (OpenVLA-best) achieves both earlier usability (SR\_$t^\ast$) and substantially higher overall sample efficiency (AUC), while PPO fails to achieve meaningful success within the training horizon.}
\label{tab:lll_success_small_merged}
\end{table}